\documentclass[11pt,a4paper]{article}
\long\def\eat#1{\ignorespaces}
\usepackage[hyperref]{naaclhlt2018}
\usepackage{times}
\usepackage{latexsym}
\usepackage{mathtools}
\usepackage{url}
\usepackage{algorithm}
\usepackage[noend]{algpseudocode}
\makeatletter
\def\BState{\State\hskip-\ALG@thistlm}
\makeatother
\usepackage{helvet}
\usepackage{courier}
\usepackage{graphicx}
\usepackage{adjustbox}
\usepackage{amsmath,amssymb,amsthm}
\usepackage{mathabx}
\usepackage{multirow}
\usepackage{url}
\usepackage{hhline}
\usepackage{enumerate}
\usepackage{pgfplots}
\usepackage{xcolor}

\usetikzlibrary{arrows,intersections}

\usetikzlibrary{pgfplots.groupplots}

\def\figlabel#1{\label{fig:#1}\label{p:#1}}
\def\tablabel#1{\label{tab:#1}\label{p:#1}}

\def\dnrm#1{\mbox{$_{\hbox{\scriptsize #1}}$}}

\def\cupequal{\cup\!\!=}

\newcommand*{\affaddr}[1]{#1} 
\newcommand*{\affmark}[1][*]{\textsuperscript{#1}}

\newcounter{notecounter}
\newcommand{\enotesoff}{\long\gdef\enote##1##2{}}

\enotesoff

\aclfinalcopy 

\title{
Joint Bootstrapping Machines for High Confidence Relation Extraction}

\author{Pankaj Gupta\affmark[1,2], Benjamin
  Roth\affmark[2], Hinrich Sch\"{u}tze\affmark[2]\\ 
 \affaddr{\affmark[1]Corporate Technology, Machine-Intelligence (MIC-DE), Siemens AG  Munich, Germany}\\
  \affaddr{\affmark[2]CIS, University of Munich (LMU) Munich, Germany} \\
  {\tt {pankaj.gupta}@siemens.com | pankaj.gupta@campus.lmu.de}\\
  {\tt  \{beroth, inquiries\}@cis.lmu.de}
}

\date{}

\eat{
\pgfplotsset{compat=1.13}
}
\def\figref#1{Figure~\ref{fig:#1}}
\def\figlabel#1{\label{fig:#1}\label{p:#1}}

\def\tabref#1{Table~\ref{tab:#1}}
\def\tablabel#1{\label{tab:#1}\label{p:#1}}

\def\eqref#1{Eq.~\ref{eqn:#1}}

\def\eqlabel#1{\label{eqn:#1}}

\begin{document}
\maketitle

\begin{abstract}

Semi-supervised bootstrapping techniques for relationship 
extraction from text iteratively expand a set of initial 
seed instances. 
Due to the lack of labeled data, a key challenge in bootstrapping is semantic drift: if a
false positive instance is added during an iteration, then
all following iterations are contaminated.
We introduce BREX,
a new bootstrapping method that protects against such
contamination by highly effective confidence assessment.
This is achieved by using entity and template seeds jointly (as
opposed to just one as in previous work),
by expanding entities and templates in parallel and in a
mutually constraining fashion in each iteration
and by introducing
higher-quality similarity measures for templates.
Experimental
results show that BREX
achieves an $F_1$ that is
0.13 (0.87 vs.\ 0.74) better
than the state of the art for four relationships.

\end{abstract}


\section{Introduction}\label{sec:intro}


Traditional semi-supervised bootstrapping relation
extractors (REs)
such as BREDS \cite{bat:82}, SnowBall \cite{gra:82} and DIPRE \cite{bri:82} require an initial set of
seed {\it entity pairs} for the target binary relation.
They find occurrences of positive seed entity pairs in the corpus, 
which are converted into extraction patterns, i.e., {\it
  extractors},
where we define an extractor as a
cluster of instances generated from the corpus.
The initial seed entity pair set  is expanded with the 
relationship entity pairs newly extracted by the extractors from the text iteratively. 
The augmented set  is then used to extract new relationships
until a stopping criterion is met.


Due to lack of sufficient labeled data, rule-based systems dominate commercial use \cite{chi:82}. Rules are typically defined by creating patterns around the entities (entity extraction) or entity pairs (relation extraction). Recently, supervised machine learning, especially deep learning techniques \cite{pkj:83, ralph:82, vu:82, Thang:81, pkj:82}, have shown promising results in entity and relation extraction; however, they need sufficient hand-labeled data to train models, which can be costly and time consuming for web-scale 
extractions. Bootstrapping machine-learned rules can make extractions easier on large corpora. Thus, open information extraction systems \cite{andrew:83, far:82, mausam:82, mes:82, ang:82} have recently been popular for domain specific or independent pattern learning.

\newcite{hea:82} used hand written rules to generate more
rules to extract hypernym-hyponym pairs, without
distributional similarity.  For entity extraction,
\newcite{ril:82} used seed entities to generate extractors
with heuristic rules and scored them by counting positive
extractions. Prior work \cite{lin:82, gupta:82}
investigated different extractor scoring measures.
\newcite{son:82} improved scores by introducing expected
number of negative entities.



\newcite{bri:82} developed the  bootstrapping relation
extraction system DIPRE that generates extractors by
clustering contexts based on string matching.  
SnowBall \cite{gra:82} is inspired by DIPRE but
computes a TF-IDF representation of each context.
BREDS \cite{bat:82} uses word embeddings \cite{mik:81} to
bootstrap relationships.






Related work investigated adapting extractor scoring
measures in bootstrapping entity extraction with either
entities or {\it templates} (\tabref{defterm}) as seeds
(Table \ref{seed-patterns}).  The state-of-the-art relation
extractors bootstrap with only seed entity pairs and suffer
due to a surplus of unknown extractions and the lack of
labeled data, leading to low confidence extractors.  This in
turn leads to 
to low confidence in the system output.
Prior RE systems
do not focus on improving the extractors' scores.  In
addition, SnowBall and BREDS used a weighting scheme to
incorporate the importance of contexts around entities and
compute a similarity score that introduces additional
parameters and does not generalize well.


\begin{table}[t]
\centering
\def\arraystretch{1.15}
\resizebox{0.48\textwidth}{!}{
\begin{tabular}{ll}
  \hline
   BREE   &  Bootstrapping Relation Extractor with {\it Entity pair}\\
   BRET   &  Bootstrapping Relation Extractor with {\it Template}\\
   BREJ   &  Bootstrapping Relation Extractor in Joint learning\\ 
   type & a named entity type, e.g., \texttt{person}\\
  typed entity & a typed entity, e.g., $<$``Obama'',\texttt{person}$>$\\
  entity pair & a pair of two typed entities\\
  template  & a triple of vectors ($\vec{v}_{-1}$, $\vec{v}_{0}$, $\vec{v}_{1}$)  and an entity pair\\  
  instance & entity pair and template (types must be the same)\\
  $\gamma$ & instance set extracted from  corpus\\
  $i$ & a member of $\gamma$, i.e., an instance\\
  $x(i)$ & the entity pair of instance $i$\\
  $\mathfrak{x}(i)$ & the template of instance $i$\\
  $G_p$ & a set of positive seed entity pairs\\
  $G_n$ & a set of negative seed entity pairs\\
  $\mathfrak{G}_p$ & a set of positive seed templates\\
  $\mathfrak{G}_n$ & a set of negative seed templates\\
${\cal G}$ & $<G_p,G_n,\mathfrak{G}_p,\mathfrak{G}_n>$\\
$k\dnrm{it}$ & number of iterations\\
$\lambda_{cat}$ & cluster of instances ({\it extractor})\\
$cat$ & category of {\it extractor} $\lambda$\\ 
$\lambda_{NNHC}$ &  Non-Noisy-High-Confidence extractor (True Positive)\\
$\lambda_{NNLC}$ & Non-Noisy-Low-Confidence extractor (True Negative)\\
$\lambda_{NHC}$ &  Noisy-High-Confidence extractor  (False Positive)\\
$\lambda_{NLC}$ & Noisy-Low-Confidence extractor  (False Negative)\\ \hline
\end{tabular}}
\caption{Notation and definition of key terms\tablabel{defterm}}
\end{table}


{\bf Contributions.} 
{\bf (1)} We propose  a {\it Joint Bootstrapping Machine}\footnote{github.com/pgcool/Joint-Bootstrapping-Machines} 
(JBM), an alternative to the entity-pair-centered bootstrapping for relation extraction  
that can take advantage of both entity-pair and template-centered methods to jointly learn extractors  
consisting of instances due to the occurrences of both entity pair and template seeds. 
It scales up the number of positive extractions for {\it non-noisy} extractors and boosts their confidence scores. 
We focus on improving the scores for {\it non-noisy-low-confidence} extractors, resulting in higher {\it recall}. 
The relation extractors bootstrapped with entity pair,
template and joint seeds are named as {\it BREE}, {\it BRET}
and {\it BREJ} (\tabref{defterm}), respectively.


{\bf (2)} Prior work on embedding-based context comparison has assumed that relations have 
\emph{consistent syntactic expression} and has mainly addressed synonymy by
 using 
embeddings (e.g.,``acquired" -- ``bought"). In reality, there is \emph{large 
variation in the syntax} of how relations are expressed, e.g., ``MSFT to acquire 
NOK for \$8B" vs.\ ``MSFT earnings hurt by NOK acquisition". We introduce 
cross-context similarities that compare all parts of the context
(e.g., ``to acquire" and ``acquisition") and show that these perform 
better (in terms of recall) than measures assuming consistent syntactic expression of relations.


{\bf (3)} Experimental results demonstrate a 13\% gain in $F1$ score on average for four relationships and suggest eliminating four parameters, 
compared to the state-of-the-art method.


The {\it motivation} and {\it benefits} of the proposed JBM for relation extraction is discussed in depth in section~\ref{sec:differences}. 
The method is applicable for both entity and relation extraction tasks. 
However, in {\it context of relation extraction}, we call it {\it BREJ}. 

\section{Method}

\subsection{Notation and definitions}
We first introduce the notation and terms (\tabref{defterm}).

Given a relationship like ``$x$ acquires $y$'',
the task is to extract pairs of entities from a corpus for which the
relationship is true. We assume that the arguments of the
relationship are typed, e.g., $x$ and $y$ are organizations.
We
run a named entity tagger in preprocessing, so that the
types of all candidate entities are given. The objects the bootstrapping
algorithm generally handles are therefore \emph{typed
  entities} (an entity associated with a type).


\enote{pkj}{current experimental results are only ordered pairs similar to BREDS.}

For a particular sentence in a corpus that states that the
relationship (e.g., ``acquires'') holds between $x$ and $y$, 
a \emph{template} consists of three vectors that represent the
context of $x$ and $y$.
$\vec{v}_{-1}$ represents the context before $x$,
$\vec{v}_{0}$  the context between $x$ and $y$ and
$\vec{v}_{1}$  the context after $y$.
These vectors are simply sums of the
embeddings of the corresponding words. 
A template is ``typed'', i.e.,
in addition to the three vectors it specifies the types of
the two entities.
An \emph{instance} 
joins an entity pair and a template. The types
of entity pair and template must be the same.

The first step of bootstrapping is to extract a set of
instances from the input corpus. We refer to this set as
$\gamma$. We will use $i$ and $j$ to refer to
instances. $x(i)$ is the entity pair of instance $i$ and 
$\mathfrak{x(i)}$ is the template of instance $i$.

A required input to our algorithm are sets of positive and negative seeds for either entity
pairs  ($G_p$ and $G_n$) 
or templates ($\mathfrak{G}_p$ and $\mathfrak{G}_n$) or both. We define $\cal G$ to be a
tuple of all four seed sets.
\enote{pkj}{$\cal G$ to be a structure, instead of tuples or vice-versa. 
We should use a common data type name. 
Later in section 1.2  para 2, you  mentioned $\cal G\dnrm{seed}$ as structure.
}

We run our bootstrapping algorithm for
$k\dnrm{it}$ iterations where
$k\dnrm{it}$ is a parameter.

A key notion is the similarity between two
instances. We will experiment with different similarity
measures. The baseline is
 \cite{bat:82}'s measure given in \figref{simmeas}, first
 line: the similarity of two instances is given as a
 weighted sum of the
 dot
 products of their before contexts ($\vec{v}_{-1}$), their
 between contexts ($\vec{v}_{0}$) and
 their after contexts ($\vec{v}_{1}$) where the weights $w_p$ are
 parameters.
We give this definition for instances, but it also applies to
templates since only the context vectors of an instance are used, not the
entities.

The similarity between an instance $i$ and a cluster
$\lambda$ of
instances is defined as the maximum similarity of $i$ with
any member of the cluster; see \figref{brexalgo}, right, \eqref{clustersim}. Again, there is a straightforward
extension to a cluster of templates: see \figref{brexalgo},
right, \eqref{clustersimtemp}.

The extractors $\Lambda$ can be categorized as follows: 
\begin{align}
\small
\begin{split}
\Lambda_{NNHC} & = \{ \lambda \in \Lambda |  \underbrace{\lambda  \mapsto \mathfrak{R}}_{non-noisy} \land \ \mbox{cnf}(\lambda, {\cal G}) \ge \tau_{cnf}\}
\end{split}\\
\small
\begin{split}
\Lambda_{NNLC} & = \{ \lambda \in \Lambda |   \lambda  \mapsto \mathfrak{R} \land \mbox{cnf}(\lambda, {\cal G}) < \tau_{cnf} \} 
\end{split}\\
\small
\begin{split}
\Lambda_{NHC} & = \{ \lambda \in \Lambda |  \underbrace{\lambda \not \mapsto \mathfrak{R}}_{noisy} \land \ \mbox{cnf}(\lambda, {\cal G}) \ge \tau_{cnf} \} 
\end{split}\\
\small
\begin{split}
\Lambda_{NLC}  & = \{ \lambda \in \Lambda |  \lambda \not \mapsto \mathfrak{R} \land \ \mbox{cnf}(\lambda, {\cal G}) < \tau_{cnf} \} 
\end{split}
\end{align}
where $\mathfrak{R}$ is the relation to be bootstrapped. The $\lambda_{cat}$ is a member of $\Lambda_{cat}$.
For instance, a $\lambda_{NNLC}$ is called as a {\it non-noisy-low-confidence} extractor if it represents the target relation (i.e., $\lambda  \mapsto \mathfrak{R}$), 
however with the confidence below a certain  threshold ($\tau_{cnf}$).   
Extractors of types $\Lambda_{NNHC}$ and $\Lambda_{NLC}$ are
desirable, those of types $\Lambda_{NHC}$ and $\Lambda_{NNLC}$ undesirable within bootstrapping. 

\subsection{ The Bootstrapping Machines: BREX}\label{BREX}
\begin{figure}[t]
 \centering 
 \includegraphics[scale=0.91]{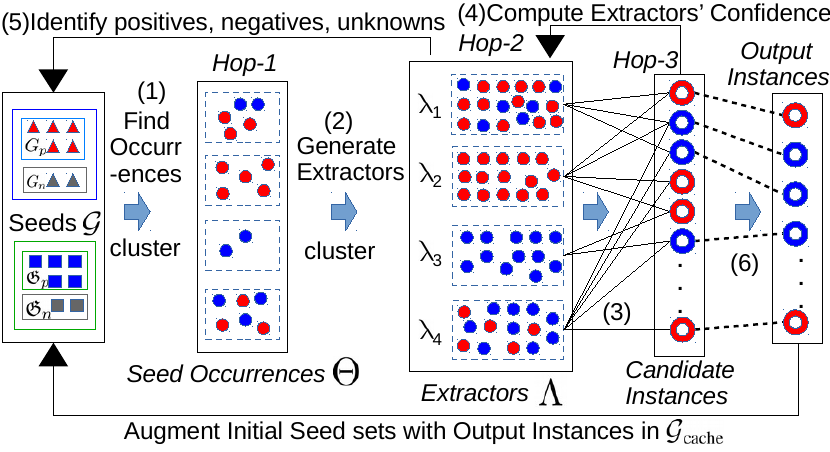}
  \caption{Joint Bootstrapping Machine. 
The red and blue filled circles/rings are the instances generated 
due to seed entity pairs and templates, respectively. 
Each dashed rectangular box represents a cluster of instances. Numbers indicate the flow. 
Follow the notations from Table 1 and Figure 2.}
  \label{fig:definitions}
\end{figure}

To describe BREX (Figure~\ref{fig:definitions}) in its most general form, we use the term
\emph{item} to refer to an entity pair, a template or both. 

The input to BREX (\figref{brexalgo}, left, line 01) is a set $\gamma$ of instances
extracted from a corpus and $\cal G\dnrm{seed}$, a structure
consisting of one set of
positive and one set of negative seed items.  $\cal G\dnrm{yield}$
(line 02) collects the items that BREX extracts in several
iterations. In each
of $k\dnrm{it}$
iterations (line 03), BREX first
initializes the cache $\cal G\dnrm{cache}$ (line 04); this cache
collects the items that are extracted in this iteration.
The design of the algorithm balances elements that ensure high recall
with elements that ensure high precision.

High recall is
achieved by starting with the seeds and making three
``hops'' that consecutively consider order-1, order-2 and order-3 neighbors
of the seeds. On line 05, we make the first hop: all
instances that are similar to a seed are
collected where ``similarity'' is defined differently for
different BREX configurations (see below).  
The collected instances are then
clustered, similar to work on bootstrapping by
\newcite{gra:82} and \newcite{bat:82}.
On line 06, we make the second hop: all
instances that are
within $\tau\dnrm{sim}$ of a hop-1 instance are added; each
such instance is only added to one cluster, the closest
one; see definition of $\mu$: \figref{brexalgo}, \eqref{clustertransform}. On line 07, we make the third hop: we include all
instances that are within
$\tau\dnrm{sim}$ of a hop-2 instance;
see definition of $\psi$: \figref{brexalgo}, \eqref{thirdhop}.
In summary, every
instance that can be reached by three hops from a seed is
being considered at this point. 
A cluster of hop-2 instances is named as {\it extractor}. 
\enote{hs}{discuss:
The clusters of hop-0 and hop-1 instances are named as {\it seed patterns} ($\Theta$) 
and {\it extraction patterns}  ($\Lambda$), respectively.}

\begin{figure*}[t]
\centering
\def\arraystretch{1.2}
\begin{tabular}{ll}
\begin{minipage}[b]{0.4\textwidth}
$\>\>$       \, \, \, \, \, \, \,     \underline{\textbf{Algorithm:}  \textit{BREX}} \\

\begin{tabbing}
 000 \= 111 \= 222 \= 333 \= 444 \= 555 \kill
01 \> INPUT: $\gamma$, ${\cal G}\dnrm{seed}$\\
02 \> $ {\cal G}\dnrm{yield} := {\cal G}\dnrm{seed}$\\
03 \> for $k\dnrm{it}$ iterations:\\
04 \> \> ${\cal G}\dnrm{cache} := \emptyset$\\
05 \> \> $\Theta := \biguplus(\{i \in \gamma| \mbox{match}(i,{\cal G}\dnrm{yield}))$\\
06 \> \> $\Lambda := \{\mu(\theta,\Theta)|\theta \in \Theta \}$\\
07 \>  \> for each $i \in \bigcup_{\lambda \in \Lambda} \psi(\lambda)$:\\
08 \>  \> \> if $\mbox{check}(i,\Lambda,{\cal G}\dnrm{yield}):$\\
09 \>  \> \> \>   $\mbox{add}(i,{\cal G}\dnrm{cache})$\\
10 \> \> $ {\cal G}\dnrm{yield} \cupequal   {\cal G}\dnrm{cache}$\\
11 \> OUTPUT: ${\cal G}\dnrm{yield}$, $\Lambda$ 
\end{tabbing}
\end{minipage}
&
\begin{minipage}[b]{0.55\textwidth}
  \begin{eqnarray}
\eqlabel{simlambda}\mbox{sim}(i,\lambda) &=&
\mbox{max}_{i' \in \lambda} \mbox{sim}(i,i') \eqlabel{clustersim}\\
\mbox{sim}(i,\mathfrak{G}) &=&
\mbox{max}_{\mathfrak{t} \in \mathfrak{G}} \mbox{sim}(i,\mathfrak{t}) \eqlabel{clustersimtemp}\\
  \psi(\lambda) &=& \{ i \in \gamma |
\mbox{sim}(i,\lambda)\geq \tau\dnrm{sim}\}\eqlabel{thirdhop}
\\
\mu(\theta,\Theta)
&=& 
\{ i \in \gamma |
\mbox{sim}(i,\theta)=d \wedge \nonumber \\
&& d=\max_{\theta \in \Theta}
\mbox{sim}(i,\theta)\geq \tau\dnrm{sim}\}\eqlabel{clustertransform}\\
\mbox{cnf}(i,\Lambda,{\cal G}) &=&
1-\!\!\!\!\!\!\!\!\!\!\!\! \prod_{\{\lambda \in
  \Lambda|i\in \psi(\lambda)\}}\!\!\!\!\!\!\!\!\!
(1\!\!-\!\!\mbox{cnf}(i,\lambda,{\cal G}))
 \eqlabel{stringentcheck} \\
\mbox{cnf}(i,\lambda,{\cal G}) &=&
\mbox{cnf}(\lambda,{\cal G})\mbox{sim}(i,\lambda)\eqlabel{instanceconf}
\\
\mbox{cnf}(\lambda,{\cal G}) &=&
\frac{
1
}{
1
+w_n\frac{N_+(\lambda,{\cal G}_n)}{N_+(\lambda,{\cal G}_p)}
+w_u\frac{N_0(\lambda,{\cal G})}{N_+(\lambda,{\cal G}_p)}
}\eqlabel{clusterconf}\\
N_0(\lambda,{\cal G}) &=& |\{ i \in \lambda |
x(i) \not\in (G_p \cup G_n)\}|  
\end{eqnarray}
\end{minipage}
\end{tabular}
  \caption{BREX algorithm (left) and definition of key
    concepts (right)\figlabel{brexalgo}}
\end{figure*}

\begin{figure*}[t]
\centering
\def\arraystretch{1.2}
\resizebox{0.99\textwidth}{!}{
  \begin{tabular}{l@{\hspace{0.02cm}}l||l|l|l} 
    && \multicolumn{1}{c}{\bf BREE} & \multicolumn{1}{c}{\bf BRET} & \multicolumn{1}{c}{\bf BREJ}\\ \hline
    & \multicolumn{1}{c||}{\it Seed Type} & \multicolumn{1}{c|}{\it Entity pairs} & \multicolumn{1}{c|}{\it Templates} & \multicolumn{1}{c}{\it Joint (Entity pairs + Templates)}\\\hline\hline
    (i)    & $N_+(\lambda,{\cal G}_l)$&
    $|\{ i \!\!\in\!\! \lambda |
x(i) \!\!\in\!\! G_l \}|$&$|
\{ i \!\!\in\!\! \lambda |
\mbox{sim}(i,\mathfrak{G}_l) \!\geq\! \tau\!\dnrm{sim} \}|$&
    $|\{ i \!\!\in\!\! \lambda |
x(i) \!\!\in\!\! G_l \}|\!+\!|
\{ i \!\!\in\!\! \lambda |
\mbox{sim}(i,\mathfrak{G}_l) \!\geq\! \tau\!\dnrm{sim} \}|\!\!$\\
(ii)&$(w_n,w_u)$ & $(1.0 ,0.0)$ & $(1.0, 0.0)$ & $(1.0,0.0)$\\
    05 &$\mbox{match}(i, {\cal G})$    & $x(i) \in
G_p$&
$\mbox{sim}(i,\mathfrak{G}_p) \!\geq\! \tau\!\dnrm{sim}$
&
 $x(i) \in  
G_p \vee
\mbox{sim}(i,\mathfrak{G}_p) \!\geq\! \tau\!\dnrm{sim}$
\\
08 & $\mbox{check}(i,\Lambda,{\cal G})$&
$\mbox{cnf}(i,\Lambda,{\cal G})\!\geq\! \tau\!\dnrm{cnf}$&
$\mbox{cnf}(i,\Lambda,{\cal G})\!\geq\! \tau\!\dnrm{cnf}$&
$\mbox{cnf}(i,\Lambda,{\cal G})\!\geq\! \tau\!\dnrm{cnf}\wedge \mbox{sim}(i,\mathfrak{G}_p) \!\geq\! \tau\!\dnrm{sim}$\\
09 & $\mbox{add}(i,{\cal G})$&
   $G_p \cupequal  \{x(i) \}$ 
&   $\mathfrak{G}_p \cupequal  \{ \mathfrak{x}(i) \}$
&   $G_p \cupequal \{x(i) \}$,
  $\mathfrak{G}_p \cupequal   \{\mathfrak{x}(i)\}$
\end{tabular}}
\caption{BREX configurations\figlabel{brexconfig}}
  \end{figure*}

High precision is achieved by imposing, on line 08, a stringent check on
each instance before its information is added to the cache.
The core function of this check is given in \figref{brexalgo}, \eqref{stringentcheck}.
This definition  is a soft
version of the following hard max, which is easier to explain: 

$ \ \ \ \mbox{cnf}(i,\Lambda,{\cal G}) \approx
 \max_{\{\lambda \in
  \Lambda|i\in \psi(\lambda)\}}
\mbox{cnf}(i,\lambda,{\cal G}) $

We are looking for a cluster
$\lambda$ in $\Lambda$ that licenses the extraction of $i$
with high confidence.
$\mbox{cnf}(i,\lambda,{\cal G})$
(\figref{brexalgo}, \eqref{instanceconf}), the {\it confidence} of
a single cluster (i.e., extractor) $\lambda$ for an instance,
is  defined as
the product of the
overall reliability of $\lambda$ (which is independent of $i$)
and the similarity of $i$ to $\lambda$, the second factor
in
\eqref{instanceconf}, i.e., $\mbox{sim}(i,\lambda)$. This factor
$\mbox{sim}(i,\lambda)$ prevents an extraction by a cluster
whose members are all distant from the instance
-- even if the cluster itself is
highly reliable.

The first factor in
\eqref{instanceconf}, i.e.,
$\mbox{cnf}(\lambda,{\cal G})$, 
assesses the reliability of a cluster $\lambda$: we
compute the ratio
$\frac{N_+(\lambda,{\cal G}_n)}{N_+(\lambda,{\cal G}_p)}$, i.e., the ratio between
the number of instances in $\lambda$ that match a negative and 
positive gold seed, respectively;
see
\figref{brexconfig}, line (i). If this ratio is close to zero, then  likely false
positive
extractions are few compared to likely true positive
extractions. For the simple version of the algorithm (for
which we set $w_n=1$, $w_u=0$), this
results in $\mbox{cnf}(\lambda,{\cal G})$ being close to 1
and the reliability measure it not discounted. 
On the other hand,  
if $\frac{N_+(\lambda,{\cal G}_n)}{N_+(\lambda,{\cal G}_p)}$ is larger, meaning that the relative number of
likely false positive extractions is high, then 
 $\mbox{cnf}(\lambda,{\cal G})$ shrinks towards 0,
resulting in progressive discounting of 
$\mbox{cnf}(\lambda,{\cal G})$ 
and leading to {\it non-noisy-low-confidence} extractor, particularly for a reliable $\lambda$.  
Due to lack of labeled data, the scoring mechanism cannot distinguish between noisy and non-noisy extractors.
Therefore, an extractor is judged by its ability to extract more positive and less negative extractions. 
Note that we carefully
designed this precision component to give good assessments
while at the same time making maximum use of the available
seeds. The reliability statistics are computed on $\lambda$,
i.e., on hop-2 instances (not on hop-3 instances). 
The ratio $\frac{N_+(\lambda,{\cal G}_n)}{N_+(\lambda,{\cal G}_p)}$ is computed on instances that directly match a gold
seed -- this is the most reliable information we have available. 

After all instances have been checked (line 08) and (if they
passed muster) added to the cache (line 09), the inner loop ends and
the cache is merged into the yield (line 10). Then a new
loop (lines 03--10) of
hop-1, hop-2 and hop-3 extensions and cluster reliability
tests starts.

\begin{figure*}[t]
\small
  \begin{eqnarray}
    \label{eq:simccv1}
\mbox{$\mbox{sim}\dnrm{match}(i,j) =  \sum_{p \in \{-1,0,1\}} w_p\vec{v}_p(i) 
\vec{v}_p(j) \ \ \ \  \ ; \  \ \ \ \  \
\mbox{sim}_{cc}^{asym} (i, j) = \max_{p \in \{-1,0,1\}}\vec{v}_p(i) 
\vec{v}_0(j)$}
\\
\label{eq:simccv2}
\mbox{$\mbox{sim}_{cc}^{sym1} (i, j)  = \max\big( \max_{p \in 
\{-1,0,1\}}\vec{v}_p(i) \vec{v}_0(j),  \max_{p \in \{-\
1,0,1\}}\vec{v}_p(j) \vec{v}_0(i) \big)$}
  \\
\mbox{$\mbox{sim}_{cc}^{sym2}  (i, j)= \max \Big( 
\big(\vec{v}_{-1}(i)+\vec{v}_1(i) \big) \vec{v}_0(j), \big( \vec{\
v}_{-1}(j) +\vec{v}_1(j) \big) \vec{v}_0(i), \vec{v}_0(i)\vec{v}_0(j) \Big)$}
\label{eq:simccv3}
\end{eqnarray}
\caption{Similarity measures.\figlabel{simmeas}
These definitions for instances equally apply to templates
since the definitions only depend on the ``template part''
of an instance, i.e., its
vectors. (value is 0 if types are different)}
\end{figure*}

Thus, the algorithm consists of $k\dnrm{it}$
iterations. There is a tradeoff
here between $\tau\dnrm{sim}$ and $k\dnrm{it}$.
We will give two extreme examples, 
assuming that we want to
extract a fixed number of $m$ instances where $m$ is given. We can achieve this goal
either 
by setting $k\dnrm{it}$=1 and
choosing a small
$\tau\dnrm{sim}$, which will result in very large hops. Or
we can achieve this goal by setting 
$\tau\dnrm{sim}$ to a large value and running the algorithm
for a larger
number of $k\dnrm{it}$.
The flexibility that the
two hyperparameters
$k\dnrm{it}$ and
$\tau\dnrm{sim}$ afford is important for good performance.

\begin{figure}[t]
 \centering 
 \includegraphics[scale=.74]{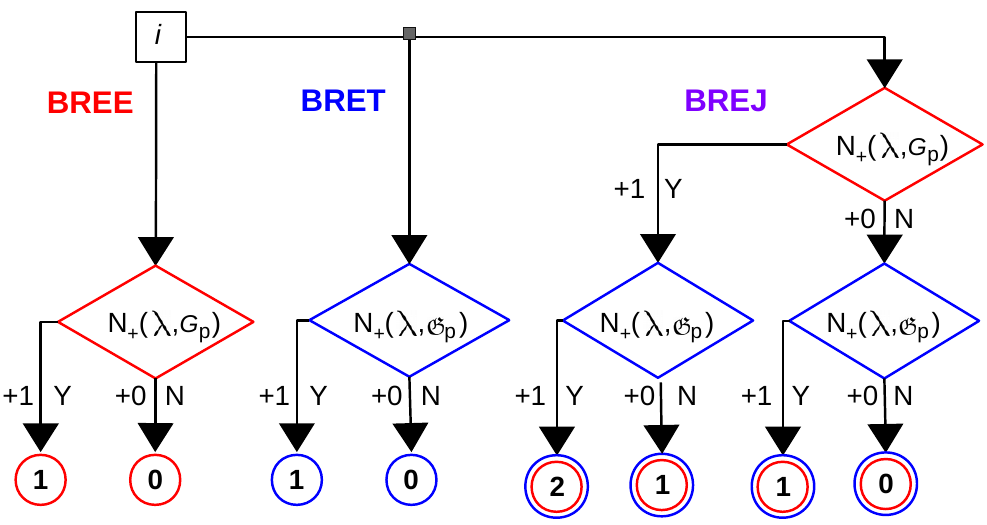}
  \caption{Illustration of Scaling-up Positive Instances. $i$: an instance in extractor, $\lambda$. Y: YES and N: NO}
  \label{fig:flowchart}
\end{figure}

\begin{figure}[t]
 \centering 
 \includegraphics[scale=.77]{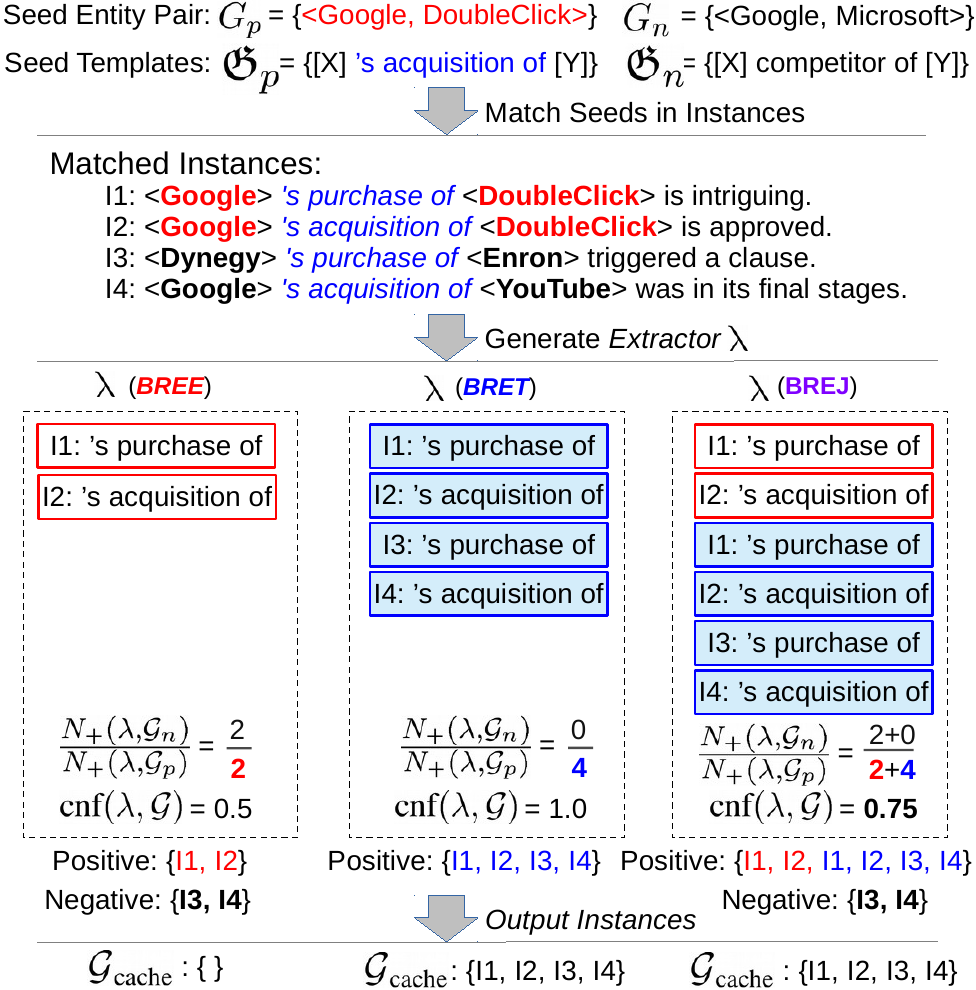}
  \caption{An illustration of scaling positive extractions and computing confidence for a non-noisy extractor generated for {\it acquired} relation. 
The dashed rectangular box represents an extractor $\lambda$, where $\lambda$ (BREJ) is {\it hybrid} with 6 instances. 
Text segments matched with seed template are shown in italics. 
Unknowns (bold in black) are considered as negatives. ${\cal G}_{cache}$ is a set of output instances where $\tau_{cnf} = 0.70$. 
}
  \label{fig:confidencescores}
\end{figure}

\subsection{BREE, BRET and BREJ}\label{sec:differences}
The main contribution of this paper is that we propose, as
an alternative to entity-pair-centered BREE \cite{bat:82},
template-centered BRET as well as BREJ (Figure \ref{fig:definitions}), an instantiation of
BREX that can take advantage of both entity pairs and
templates. The differences and advantages of BREJ over BREE and BRET are:

{\bf (1) Disjunctive Matching of Instances:} 
The first difference is realized in how
the three algorithms match instances with seeds (line 05
in \figref{brexconfig}).
BREE checks whether the entity pair of an instance is one
of the entity pair seeds, BRET checks 
whether the template of an instance is one
of the template seeds and BREJ checks whether the
disjunction of the two is true. 
The disjunction facilitates a higher hit rate in matching instances  with seeds. 
The introduction of a few handcrafted templates along with seed entity pairs allows BREJ to 
leverage discriminative patterns and learn similar ones via distributional semantics.    
In Figure \ref{fig:definitions}, the joint approach results in {\it hybrid} extractors $\Lambda$ that contain instances 
due to seed occurrences $\Theta$ of both entity pairs and templates.  

{\bf (2) Hybrid Augmentation of Seeds:} 
On line 09 in \figref{brexconfig},
we see that the bootstrapping  step is defined in a
straightforward fashion: the entity pair of an instance is added for BREE,
the template for BRET and both for BREJ. 
Figure \ref{fig:definitions} demonstrates the hybrid augmentation of seeds via {\it red} and {\it blue} rings of {\it output instances}.   

\begin{table*}[t]
\centering
\resizebox{0.98\textwidth}{!}{
\def\arraystretch{1.2}
\begin{tabular}{l|cc}
\multicolumn{1}{l}{\multirow{1}{*}{}{\it {Relationship}}}                             & {\it Seed Entity Pairs}                                    & {\it Seed Templates}         \\                                                                                                                                                                                         
\hline
acquired                                          & \begin{tabular}[c]{@{}c@{}}\{Adidas;Reebok\},\{Google;DoubleClick\},\\ \{Widnes;Warrington\},\{Hewlett-Packard;Compaq\}\end{tabular}      & \begin{tabular}[c]{@{}c@{}}\{{[}X{]} acquire {[}Y{]}\},\{{[}X{]} acquisition {[}Y{]}\},\{{[}X{]} buy {[}Y{]}\},\\ \{{[}X{]} takeover {[}Y{]}\},\{{[}X{]} merger with {[}Y{]}\}\end{tabular}              \\
\hline
founder-of                                        & \begin{tabular}[c]{@{}c@{}}\{CNN;Ted Turner\},\{Facebook;Mark Zuckerberg\},\\ \{Microsoft;Paul Allen\},\{Amazon;Jeff Bezos\},\end{tabular} & \begin{tabular}[c]{@{}c@{}}\{{[}X{]} founded by {[}Y{]}\},\{{[}X{]} co-founder {[}Y{]}\},\{{[}X{]} started by {[}Y{]}\},\\ \{{[}X{]} founder of {[}Y{]}\},\{{[}X{]} owner of {[}Y{]}\}\end{tabular}  \\
\hline
headquartered                                     & \begin{tabular}[c]{@{}c@{}}\{Nokia;Espoo\},\{Pfizer;New York\},\\ \{United Nations;New York\},\{NATO;Brussels\},\end{tabular}         & \begin{tabular}[c]{@{}c@{}}\{{[}X{]} based in {[}Y{]}\},\{{[}X{]} headquarters in {[}Y{]}\},\{{[}X{]} head office in {[}Y{]}\},\\ \{{[}X{]} main office building in {[}Y{]}\},\{{[}X{]} campus branch in {[}Y{]}\}\end{tabular}                                                                                                                                                                                        \\
\hline
affiliation                                       & \begin{tabular}[c]{@{}c@{}}\{Google;Marissa Mayer\},\{Xerox;Ursula Burns\},\\ \{ Microsoft;Steve Ballmer\},\{Microsoft;Bill Gates\},\end{tabular}      & \begin{tabular}[c]{@{}c@{}}\{{[}X{]} CEO {[}Y{]}\},\{{[}X{]} resign from {[}Y{]}\},\{{[}X{]} founded by {[}Y{]}\},\\ \{{[}X{]} worked for {[}Y{]}\},\{{[}X{]} chairman director {[}Y{]}\}\end{tabular}                                                                                                                                                                                                                                                
\end{tabular}
}
\caption{Seed Entity Pairs and Templates for each relation. [X] and  [Y] are slots for entity type tags.}
\label{seed-patterns}
\end{table*}

{\bf (3) Scaling Up Positives in Extractors:}
As discussed in section \ref{BREX}, a good measure of the quality of an 
extractor is crucial and $N_+$, the number of
instances in an extractor $\lambda$ that match a seed,
is an important
component of that. For BREE and BRET, the definition follows directly
from the fact that these are entity-pair and template-centered instantiations
of BREX, respectively. 
However, the disjunctive matching of instances for an extractor with entity pair and template seeds in BREJ (\figref{brexconfig} line ``(i)'' )
boosts the likelihood of finding positive instances. 
In Figure \ref{fig:flowchart}, we demonstrate computing the count of positive instances $N_{+}(\lambda, {\cal G})$ 
for an extractor $\lambda$ within the three systems.  Observe that an instance $i$ in $\lambda$ can scale its $N_{+}(\lambda, {\cal G})$
by a factor of maximum 2 in BREJ if $i$ is matched in both entity pair and template seeds. 
The reliability $\mbox{cnf}(\lambda, {\cal G})$ (\eqref{clusterconf}) of an extractor $\lambda$ 
is based on the ratio $\frac{N_+(\lambda,{\cal G}_n)}{N_+(\lambda,{\cal G}_p)}$, therefore suggesting that 
the scaling 
boosts its confidence. 

In Figure \ref{fig:confidencescores}, we demonstrate with an example how the joint bootstrapping scales up 
the positive instances for a {\it non-noisy} extractor $\lambda$, resulting in $\lambda_{NNHC}$ for BREJ compared to $\lambda_{NNLC}$ in BREE.  

Due to unlabeled data, the instances not matching in seeds 
are considered either to be ignored/unknown $N_{0}$ or negatives in the confidence measure (\eqref{clusterconf}). 
The former leads to high confidences for noisy extractors by assigning high scores, 
the latter to low confidences for non-noisy extractors by penalizing them. 
For a simple version of the algorithm in the illustration, we consider them as negatives and set $w_n=1$.  
Figure~\ref{fig:confidencescores} shows the three extractors ($\lambda$) generated and their confidence scores in BREE, BRET and BREJ. 
Observe that the scaling up of positives in BREJ due to BRET extractions (without $w_{n}$) discounts $\mbox{cnf}(\lambda, {\cal G})$ relatively lower than BREE. 
The discounting results in $\lambda_{NNHC}$ in BREJ and $\lambda_{NNLC}$ in BREE. 
The discounting in BREJ is adapted for {\it non-noisy} extractors facilitated by BRET 
in generating mostly non-noisy extractors due to stringent checks (\figref{brexconfig}, line ``(i)" and 05). 
Intuitively, the intermixing of non-noisy extractors (i.e., {\it hybrid}) promotes the scaling and boosts recall. 

\subsection{Similarity Measures}\label{similarity measures}
The before ($\vec{v}_{-1}$) and after ($\vec{v}_1$) contexts around the entities 
are highly sparse 
due to large variation in the syntax of how relations are expressed. 
SnowBall, DIPRE and BREE assumed that the between ($\vec{v}_0$) context mostly defines the syntactic expression for a relation and  
used weighted mechanism on 
the three contextual similarities in pairs, $\mbox{sim}_{match}$ (\figref{simmeas}).
They assigned higher weights to the similarity in between ($p=0$) contexts, that resulted in lower recall.  
We introduce attentive ($\max$) similarity across all contexts (for example, $\vec{v}_{-1}(i) \vec{v}_{0}(j)$) 
to automatically capture the large variation in the syntax of
 how relations are expressed, without using any weights. 
We investigate asymmetric (Eq \ref{eq:simccv1})  and symmetric (Eq \ref{eq:simccv2} and \ref{eq:simccv3}) similarity 
measures, 
and name them as {\it cross-context attentive} ($\mbox{sim}_{cc}$) similarity. 

\begin{table}[]
\centering
\def\arraystretch{1.12}
\small
\resizebox{0.35\textwidth}{!}{
\begin{tabular}{c|ccc}
   & {\it ORG-ORG} & {\it ORG-PER}  &{\it ORG-LOC}  \\
\hline
{\it count} &58,500 &75,600   &95,900      
\end{tabular}}
\caption{Count of entity-type pairs in corpus}
\label{candidates}
\end{table}

\begin{table}[t]
\centering
\def\arraystretch{1.15}
\resizebox{0.48\textwidth}{!}{
\begin{tabular}{c|c|c}
 {\it Parameter} & {\it  Description/ Search} & {\it Optimal}\\
\hline
$|v_{-1}|$ &  maximum number of tokens in before context & 2 \\  
$|v_{0}|$ &  maximum number of tokens in between context & 6 \\  
$|v_{1}|$ &  maximum number of tokens in after context & 2 \\  
$\tau_{sim}$    & similarity threshold  [0.6, 0.7, 0.8] & 0.7\\
$\tau_{cnf}$    & instance confidence thresholds  [0.6, 0.7, 0.8]   & 0.7\\
$w_{n}$   &  weights to negative extractions   [0.0, 0.5, 1.0, 2.0] & 0.5\\
$w_{u}$   &   weights to unknown extractions [0.0001, 0.00001] & 0.0001\\
$k_{it}$ & number of bootstrapping epochs & 3\\
$dim_{emb}$ & dimension of embedding vector, $V$& 300\\
$PMI$ & PMI threshold in evaluation & 0.5 \\
{\it Entity Pairs} & Ordered Pairs ($OP$) or Bisets ($BS$) & $OP$ 
\end{tabular}}
\caption{Hyperparameters in BREE, BRET and BREJ}
\label{hyperparameters}
\end{table}

\begin{table*}[t]
\centering
\def\arraystretch{1.22}
\resizebox{0.99\textwidth}{!}{
\begin{tabular}{cc|cccc||cccc|cccc|cccc}
&{\it Relationships} &$\#out$ & $P$    & $R$           & $F1$      &$\#out$ & $P$    & $R$           & $F1$      &$\#out$ & $P$    & $R$           & $F1$  &$\#out$ & $P$    & $R$           & $F1$  \\
\hline
\parbox[t]{2mm}{\multirow{6}{*}{\rotatebox[origin=c]{90}{\bf BREE}}}  &            &  \multicolumn{4}{c||}{{\bf baseline}: BREE+$\mbox{sim}_{match}$}       & \multicolumn{4}{c|}{{\bf config\textsubscript{2}}: BREE+$\mbox{sim}_{cc}^{asym}$}   &  \multicolumn{4}{c|}{{\bf config\textsubscript{3}}: BREE+$\mbox{sim}_{cc}^{sym1}$}     & \multicolumn{4}{c}{{\bf config\textsubscript{4}}: BREE+$\mbox{sim}_{cc}^{sym2}$}    \\
\cline{4-5} \cline{8-9}  \cline{12-13}  \cline{16-17}
 & acquired          &2687      & 0.88  & 0.48        & 0.62       &5771      & 0.88  & \underline{0.66}        & 0.76   &    3471          &     0.88          &     \underline{0.55}        &  0.68      &        3279  &  0.88      &     \underline{0.53}     & 0.66 \\
&founder-of        &628   & 0.98 & 0.70        & 0.82   &9553        & {0.86} &  \underline{0.95 }       &   0.89       &     1532         &   0.94            &    \underline{0.84}         &  0.89      &  1182       &   0.95       &    \underline{0.81}     & 0.87 \\
&headquartered &16786      & 0.62 & 0.80      &0.69     &21299      & 0.66 &  \underline{0.85}        & 0.74        &    17301          &   0.70            &   \underline{0.83}          &   0.76     &        9842  &  0.72      &    \underline{0.74}      & 0.73 \\
&affiliation   &20948  & 0.99 & 0.73        & 0.84  &27424      & 0.97 &  \underline{0.78 }       & 0.87   &    36797          &    0.95           &    \underline{0.82}         &  0.88      &   28416      &  0.97      &    \underline{0.78}      & 0.87 \\ \cline{2-18}
& {\bf avg}   & 10262   & 0.86  & 0.68        & 0.74  &  16011    & 0.84    &  \underline{0.81 }       & \underline{0.82}   &      14475        &    0.87           &    \underline{0.76}         &  \underline{0.80}      &     10680    &  0.88      &    \underline{0.72}      &  \underline{0.78} \\ 
\hline
\parbox[t]{2mm}{\multirow{6}{*}{\rotatebox[origin=c]{90}{\bf BRET}}}  &            &  \multicolumn{4}{c||}{{\bf config\textsubscript{5}}: BRET+$\mbox{sim}_{match}$}       & \multicolumn{4}{c|}{{\bf config\textsubscript{6}}: BRET+$\mbox{sim}_{cc}^{asym}$}   &  \multicolumn{4}{c|}{{\bf config\textsubscript{7}}: BRET+$\mbox{sim}_{cc}^{sym1}$}     & \multicolumn{4}{c}{{\bf config\textsubscript{8}}: BRET+$\mbox{sim}_{cc}^{sym2}$}    \\
\cline{4-5} \cline{8-9}  \cline{12-13}  \cline{16-17}    
&acquired   &  4206         & {0.99}  &  0.62        & 0.76      & 15666       & {0.90}  & \underline{0.85}        &  {0.87}     &    18273          &    0.87           &    \underline{0.86}          &  0.87      &    14319     & 0.92       &  \underline{0.84}        & 0.87 \\
&founder-of   &920       & {0.97} & 0.77        & 0.86     & 43554       & 0.81 & \underline{0.98}        & 0.89     &    41978          &      0.81         &   \underline{0.99}         &  0.89      &     46453    &    0.81    &    \underline{0.99}     & 0.89\\
&headquartered &3065      & {0.98} & 0.55        & 0.72    &39267      & 0.68 & \underline{0.92}        &  0.78            &    36374          &      0.71         &    \underline{0.91}          &       0.80  &  56815       & 0.69       &   \underline{0.94}       & {0.80} \\
&affiliation  &20726       & {0.99} & 0.73       & 0.85   &  28822       & {0.99} & \underline{0.79}       &  0.88    &      44946        &    0.96           &    \underline{0.85}          &   0.90     &    33938     &    0.97    &    \underline{0.81}      & 0.89\\  \cline{2-18}
& {\bf avg}   & 7229   & {0.98}    & 0.67        & 0.80  & 31827     & 0.85 &  \underline{0.89}       & \underline{0.86}   &       35393       &    0.84           &    \underline{0.90}         &  \underline{0.86}      &     37881    &  0.85      &    \underline{0.90}      &  \underline{0.86}\\
\hline
\parbox[t]{2mm}{\multirow{6}{*}{\rotatebox[origin=c]{90}{\bf BREJ}}}  &            &  \multicolumn{4}{c||}{{\bf config\textsubscript{9}}: BREJ+$\mbox{sim}_{match}$}       & \multicolumn{4}{c|}{{\bf config\textsubscript{10}}: BREJ+$\mbox{sim}_{cc}^{asym}$}   &  \multicolumn{4}{c|}{{\bf config\textsubscript{11}}: BREJ+$\mbox{sim}_{cc}^{sym1}$}     & \multicolumn{4}{c}{{\bf config\textsubscript{12}}: BREJ+$\mbox{sim}_{cc}^{sym2}$}    \\
\cline{4-5} \cline{8-9}  \cline{12-13}  \cline{16-17}
 &acquired      &20186    & 0.82 & {\bf 0.87}        & {\bf 0.84}   &  {35553}      & 0.80  & \underline{\bf 0.92}      & 0.86  &   22975           &      0.86         &   \underline{0.89}          &   0.87     &     22808   &        0.85 &    \underline{0.90}     &    \underline{\bf 0.88}\\
& founder-of   & 45005       & 0.81 & {\bf 0.99}        & {\bf 0.89}   & {57710}      & 0.81 & \underline{\bf 1.00}       & \underline{\bf 0.90}  &     50237         &      0.81         &    \underline{0.99}      & 0.89       &     45374   &        0.82 &    \underline{0.99}     &    0.90     \\ 
& headquartered & 47010      & 0.64 & {\bf 0.93}        & {\bf 0.76}     &  {66563}   & {0.68}  & \underline{\bf 0.96}        & \underline{\bf 0.80}    &      60495        &     0.68          &     \underline{0.94}         &    0.79    &     57853    &    0.68    &    \underline{0.94}      & 0.79 \\
& affiliation   & 40959      & 0.96   & {\bf 0.84}        & {\bf 0.89}    & {57301}       & 0.94 &   \underline{\bf 0.88}      & \underline{\bf 0.91}   &      55811        &      0.94         &    \underline{0.87}          &         {0.91}    &   51638  &  0.94      &    \underline{0.87}      & {0.90} \\ \cline{2-18} 
& {\bf avg}   & {\bf 38290}    & 0.81     & {\bf 0.91}        &  {\bf 0.85}  &  \underline{\bf 54282}    & 0.81   &  \underline{\bf 0.94 }       & \underline{\bf 0.87}   &     47380         &    0.82           &     \underline{0.92}         &  \underline{0.87}      &   44418      &  0.82      &    \underline{0.93}      & \underline{0.87} 
\end{tabular}}
\caption{Precision ($P$), Recall ($R$) and $F1$  
compared to the state-of-the-art ({\it baseline}).
$\#out$: count of output instances with $\mbox{cnf}(i, \Lambda, {\cal G})$ $\ge$ 0.5. {\bf avg}: average. 
{\bf Bold} and \underline{underline}: Maximum due to BREJ and $\mbox{sim}\dnrm{cc}$, respectively.}
\label{BRESH-results}
\end{table*}

\section{Evaluation}

\subsection{Dataset and Experimental Setup}
We re-run BREE~\cite{bat:82} for {\bf baseline} with a set of 5.5 million news articles from AFP and APW~\cite{par:82}.
We use processed dataset of 1.2 million sentences  (released by BREE) containing at least two entities linked to FreebaseEasy~\cite{Bast:82}. 
We extract four relationships: {\it acquired} (ORG-ORG), {\it founder-of} (ORG-PER), {\it headquartered} (ORG-LOC) and {\it affiliation} (ORG-PER) for Organization (ORG), Person (PER) and Location (LOC) entity types.  
We bootstrap relations in BREE, BRET and BREJ, each with 4  similarity measures using seed entity pairs and templates (Table~\ref{seed-patterns}). 
See Tables \ref{candidates},  \ref{hyperparameters} and \ref{BRESH-results} for the count of candidates, hyperparameters and 
different configurations, respectively.

Our evaluation is based on 
\newcite{bro:82}'s framework to estimate precision and recall of large-scale RE systems using FreebaseEasy \cite{Bast:82}. 
Also following \newcite{bro:82}, we use Pointwise Mutual Information (PMI) \cite{tur:82} to evaluate our system automatically, 
in addition to relying on an external knowledge base.
We consider only extracted relationship
instances with confidence scores $\mbox{cnf}(i, \Lambda, {\cal G})$ equal or above 0.5. 
We follow the same approach as BREE \cite{bat:82} to detect  the correct order of entities in a relational triple, where we try to identify the presence of passive voice using part-of-speech (POS) tags and considering any form of the verb to be, followed by a verb in the past tense or past participle, and ending in the word `by'. 
We use GloVe~\cite{pen:82} embeddings. 


\subsection{Results and Comparison with baseline}
Table~\ref{BRESH-results} shows the experimental results in the three systems for the different relationships with {\it ordered} entity pairs and  
 similarity measures ($\mbox{sim}_{match}$,  $\mbox{sim}_{cc}$). 
Observe that BRET (config\textsubscript{5}) is {\it precision-oriented} while BREJ (config\textsubscript{9}) {\it recall-oriented} when compared to BREE (baseline).  
We see the number of output instances $\#out$ are also higher in BREJ, therefore the higher recall. 
The BREJ system in the different similarity configurations outperforms the baseline BREE and BRET in terms of $F1$ score. 
On an average for the four relations, BREJ in configurations config\textsubscript{9} and config\textsubscript{10} results in $F1$ that is $0.11$ (0.85 vs 0.74) and $0.13$ (0.87 vs 0.74) better than  
the baseline BREE.  
\begin{table}[t]
\centering
\small
\def\arraystretch{1.2}
\resizebox{0.32\textwidth}{!}{
\begin{tabular}{r|c|cccc}
       $\tau$              & $k_{it}$       & $\#out$        & $P$       & $R$        &  $F1$ \\ \hline
\multirow{2}{*}{0.6} & 1    & 691   &  0.99 &  0.21 &  0.35  \\
                     & 2    & 11288 &  0.85 &  {\bf 0.79} &  0.81  \\ \hline
\multirow{2}{*}{0.7} & 1    & 610   & 1.0  & 0.19  &   0.32 \\
                     & 2    & 7948  & {\bf 0.93}  &  0.75 &  {\bf 0.83}  \\ \hline
\multirow{2}{*}{0.8} & 1    & 522   &  1.0  & 0.17  & 0.29   \\
                     & 2    & 2969  &  0.90 & 0.51  &  0.65  
\end{tabular}}
\caption{Iterations ($k_{it}$) Vs Scores with thresholds ($\tau$) 
for relation {\it acquired} in BREJ. $\tau$ refers to $\tau_{sim}$ and $\tau_{cnf}$}
\label{itervsf1}
\end{table}
\begin{table}[t]
\centering
\def\arraystretch{1.2}
\resizebox{0.46\textwidth}{!}{
\begin{tabular}{rc|cccc||c|cccc}
           & $\tau$     & $\#out$        & $P$       & $R$        &  $F1$   & $\tau$     & $\#out$        & $P$       & $R$        &  $F1$\\ \cline{2-11}
\parbox[t]{2mm}{\multirow{2}{*}{\rotatebox[origin=c]{90}{\small  BREE}}}  
                     & .60    & 1785 & .91 &  .39 &  .55 & .70    & 1222 &  .94 &  .31 & .47   \\
                     & .80    & 868 &  .95 &  .25 & .39  &  .90    & 626 &  .96 &  .19 &  .32 \\ \cline{2-11}
                    
\parbox[t]{2mm}{\multirow{2}{*}{\rotatebox[origin=c]{90}{\small  BRET}}}    
                     & .60    & 2995 &  .89 &  .51 &  .65  & .70    & 1859 &  .90 &  .40 &  .55 \\
                    & .80    & 1312 &  .91 &  .32 &  .47   & .90    & 752 &  .94 &  .22 &  .35  \\ \cline{2-11}                   

\parbox[t]{2mm}{\multirow{2}{*}{\rotatebox[origin=c]{90}{\small  BREJ}}}    
                     & .60    & 18271 &  .81 &  .85 &  .83  & .70    & 14900 &  .84 &  .83 &  .83\\
                     & .80    & 8896 &  .88 &  .75 &  .81  & .90    & 5158 &  .93 &  .65 &  .77
\end{tabular}}
\caption{Comparative analysis using different thresholds $\tau$ to evaluate the extracted instances for {\it acquired}}
\label{differentPMI}
\end{table}

We discover that $\mbox{sim}_{cc}$ improves $\#out$ and {\it recall} over $\mbox{sim}\dnrm{match}$ correspondingly in all three systems.
Observe that $\mbox{sim}_{cc}$ performs better with BRET than BREE due to {\it non-noisy} extractors in BRET. 
The results suggest an alternative to the weighting scheme in $\mbox{sim}\dnrm{match}$ and 
therefore, the state-of-the-art ($\mbox{sim}_{cc}$) performance with the 3 parameters ($w_{-1}$, $w_{0}$ and $w_{1}$) ignored in bootstrapping. 
Observe that $\mbox{sim}_{cc}^{asym}$ gives higher recall than the two symmetric similarity measures.

Table \ref{itervsf1} shows the performance of BREJ in different iterations trained with different 
similarity $\tau_{sim}$ and confidence $\tau_{cnf}$ thresholds. 
Table \ref{differentPMI} shows a comparative analysis of the three systems, 
where we consider and evaluate the extracted relationship instances 
at different confidence scores.

\subsection{Disjunctive Seed Matching of Instances}
As discussed in section~\ref{sec:differences}, BREJ facilitates disjunctive matching of instances (line 05 \figref{brexconfig}) with seed entity pairs and templates. 
Table~\ref{figdisjunctivematch} shows $\#hit$ in the three systems, where the higher values of $\#hit$ in BREJ conform to the desired property. 
Observe that some instances in BREJ are found to be matched in both the seed types.

\begin{table}[t]
\centering
\def\arraystretch{1.2}
\resizebox{0.49\textwidth}{!}{
\begin{tabular}{l|ccc|ccc|ccc|ccc}
     & \multicolumn{3}{c|}{\it acquired} & \multicolumn{3}{c|}{\it founder-of} & \multicolumn{3}{c|}{\it headquartered} & \multicolumn{3}{c}{\it affiliation} \\ \cline{2-13} 
    BRE{\bf X} & {\bf E}     & {\bf T}    & {\bf J}    & {\bf E}     & {\bf T}    & {\bf J}     & {\bf E}     & {\bf T}    & {\bf J}      & {\bf E}     & {\bf T}    & {\bf J}     \\ \hline
   $\#hit$  &   71       &   682     &  \underline{743}     &   135     &     956    &   \underline{1042}    &  715        & 3447     &  \underline{4023}      & 603       & 14888    &  \underline{15052}
\end{tabular}}
\caption{Disjunctive matching of Instances. $\#hit$: the count of instances matched to positive seeds in $k_{it}=1$}
\label{figdisjunctivematch}
\end{table}

\begin{table}[t]
\centering
\def\arraystretch{1.25}
\resizebox{0.49\textwidth}{!}{
\begin{tabular}{cc|cccccccc|c}
\multicolumn{2}{c|}{Attributes} & $|\Lambda|$      & $AIE$       & $AES$      & $ANE$     & $ANNE$  & $ANNLC$   & $AP$     & $AN$    & $ANP$\\
\hline
\parbox[t]{2mm}{\multirow{3}{*}{\rotatebox[origin=c]{90}{\small  acquired}}}  & BREE    &         167        &      12.7         &       0.51     &        0.84         &  0.16     & 0.14      &      37.7      &     93.1             &           2.46          \\
& BRET    &         17         &      305.2        &       1.00    &       0.11        &  0.89   &  0.00     &     671.8       &         0.12         &            0.00         \\
& BREJ    &         555      &      {\bf  41.6}         &       {\bf 0.74}   &       {\bf 0.71}     & {\bf  0.29}   &  {\bf 0.03}      &   {\bf 313.2}         &     {\bf 44.8}             &       {\bf 0.14}   \\
\cline{2-11}                        
\parbox[t]{2mm}{\multirow{3}{*}{\rotatebox[origin=c]{90}{\small  founder-of}}}  &  BREE   &         8        &      13.3           &       0.46        &    0.75            &     0.25  &  0.12     &   44.9        &  600.5    &      13.37         \\
& BRET     &         5        &      179.0           &       1.00       &    0.00            &     1.00  &  0.00     &   372.2        &    0.0   &      0.00       \\
& BREJ     &         492        &       {\bf 109.1}           &        {\bf 0.90}       &    0.94            &     0.06  &   {\bf 0.00 }   &    {\bf 451.8}        &   {\bf 79.5}     &      {\bf  0.18}        \\
\cline{2-11}                        
\parbox[t]{2mm}{\multirow{3}{*}{\rotatebox[origin=c]{90}{\small  headquartered}}}  &  BREE     &         655        &      18.4           &       0.60        &    0.97            &    0.03   &  0.02     &   46.3        &     82.7  &      1.78         \\
& BRET   &         7        &      365.7           &       1.00        &    0.00            &     1.00  &  0.00     &   848.6        &    0.0   &      0.00         \\
& BREJ    &         1311        &      {\bf 45.5 }          &      {\bf 0.80}       &    0.98            &     0.02  &  {\bf 0.00}     &   {\bf 324.1}        &    {\bf 77.5}   &     {\bf  0.24}         \\
\cline{2-11}                        
\parbox[t]{2mm}{\multirow{3}{*}{\rotatebox[origin=c]{90}{\small  affiliation}}}  & BREE     &         198        &      99.7           &       0.55           &    0.25            &     0.75   &  0.34     &   240.5        &  152.2     &      0.63         \\
& BRET   &         19           &    846.9         &       1.00           &        0.00      &     1.00   &    0.00   &    2137.0       &   0.0    &    0.00               \\
& BREJ    &       470    &      {\bf  130.2} &       {\bf 0.72}  &      {\bf 0.21}     &      {\bf 0.79}       & {\bf 0.06}      &    {\bf  567.6}       &   {\bf 122.7}    & {\bf  0.22}
\end{tabular}
}
\caption{Analyzing the attributes of extractors $\Lambda$ learned for each relationship. 
Attributes are: 
number of extractors ($|\Lambda|$), 
$avg$ number of instances in $\Lambda$ (AIE),  
$avg$ $\Lambda$ score (AES), 
$avg$ number of noisy $\Lambda$ (ANE),  
$avg$ number of non-noisy $\Lambda$ (ANNE),  
$avg$ number of $\Lambda_{NNLC}$ below confidence 0.5 (ANNLC), 
$avg$ number of positives (AP) and negatives (AN), 
ratio of AN to AP (ANP). 
The {\bf bold} indicates comparison of BREE and BREJ with $sim_{match}$. $avg$: average}
\label{patternscores}
\end{table}

\begin{table}[t]
\centering
\small
\def\arraystretch{1.17}
\resizebox{0.38\textwidth}{!}{
\begin{tabular}{cc|cccc}
&{\it Relationships} &$\#out$ & $P$    & $R$           & $F1$    \\
\hline
\parbox[t]{2mm}{\multirow{4}{*}{\rotatebox[origin=c]{90}{\bf BREE}}}  &acquired          & 387     & 0.99  &  0.13       &   0.23    \\
&founder-of        &  28    & 0.96  &   0.09      &     0.17   \\
&headquartered       & 672     & 0.95  &   0.21      &   0.34    \\  
&affiliation   &    17516  &  0.99 &    0.68     &    0.80  \\ \cline{2-6}
& {\bf avg}   &   4651   &  0.97 &    0.28     &    0.39  \\ \hline
\parbox[t]{2mm}{\multirow{4}{*}{\rotatebox[origin=c]{90}{\bf BRET}}} &acquired   &   4031   &  1.00  &   0.61      &  0.76   \\
&founder-of   & 920     &  0.97  &   0.77      & 0.86     \\
&headquartered & 3522     & 0.98  &  0.59       & 0.73     \\
&affiliation  &  22062     & 0.99  &     0.74    &  0.85  \\ \cline{2-6}
& {\bf avg}   &  7634    &  0.99  &    0.68     &    0.80  \\ \hline
\parbox[t]{2mm}{\multirow{4}{*}{\rotatebox[origin=c]{90}{\bf BREJ}}} &acquired     &  \underline{12278}    & 0.87  &   \underline{0.81}      & {\bf 0.84}   \\
& founder-of   &   {23727}   & 0.80  &  \underline{ 0.99}      & {\bf  0.89}     \\
& headquartered &  {38737}    & 0.61  &    \underline{0.91}     &  {\bf 0.73}   \\
& affiliation   &   {33203}   &   0.98 &   \underline{0.81}     &  {\bf 0.89} \\  \cline{2-6}
& {\bf avg}   &   {26986}   &  0.82 &    \underline{0.88}     &   {\bf 0.84} 
\end{tabular}}
\caption{BREX+sim\dnrm{match}:Scores when $w_n$ ignored}
\label{wnegtignored}
\end{table}

\begin{table*}[t]
\centering
\def\arraystretch{1.2}
\resizebox{.99\textwidth}{!}{
\begin{tabular}{cccccccc}
\bf config\textsubscript{1}: BREE + $\mbox{sim}_{\bf match}$ & \bf $\mbox{cnf}(\bf \lambda, {\cal \bf G})$ &  \bf config\textsubscript{5}: BRET + $\mbox{sim}_{\bf match}$ &  \bf $\mbox{cnf}(\bf \lambda, {\cal \bf G})$ & \bf config\textsubscript{9}: BREJ + $\mbox{sim}_{\bf match}$ &  \bf $\mbox{cnf}(\bf \lambda, {\cal \bf G})$  & \bf config\textsubscript{10}: BREJ + $\mbox{sim}_{\bf cc}^{\bf asym}$   & \bf  \bf $\mbox{cnf}(\bf \lambda, {\cal \bf G})$ \\
\hline
     \multicolumn{8}{c}{\bf acquired}\\
\hline
{[X]} acquired {[Y]}  & 0.98 & {[X]} acquired {[Y]}   & 1.00  & {[X]} acquired  {[Y]}    & 1.00  &  acquired by {[X]} ,  {[Y]}  $^{\dagger}$     & 0.93  \\
{[X]} takeover of {[Y]}    & 0.89  &    {[X]}  takeover of {[Y]}    & 1.00  & {[X]}  takeover of {[Y]}    & 0.98  &  takeover of {[X]}  would boost {[Y] 's earnings}  $^{\dagger}$   & 0.90  \\
{[X]} 's planned acquisition of {[Y]}    & 0.87  &  {[X]} 's planned acquisition of{[Y]}     & 1.00  & {[X]} 's planned acquisition of {[Y]}    & 0.98  &  acquisition of {[X]} by  {[Y]} $^{\dagger}$   & 0.95  \\
{[X]} acquiring {[Y]}    & 0.75  &    {[X]}  acquiring {[Y]}    & 1.00  & {[X]} acquiring {[Y]}    & 0.95  &    {[X]} acquiring {[Y]}    & 0.95    \\ 
{[X]}  has owned part of {[Y]} & 0.67 &  {[X]}  has owned part of {[Y]}    & 1.00  & {[X]} has owned part of {[Y]}    & 0.88  &    owned by {[X]} 's parent   {[Y]}   & 0.90   \\
{[X]}  took control of   {[Y]}   & 0.49 &  {[X]}  's ownership of {[Y]}    & 1.00  & {[X]} took control of {[Y]}    & 0.91  &   {[X]} takes control of  {[Y]}    & 1.00    \\ 
{[X]} 's acquisition of {[Y]}    & 0.35  &    {[X]}  's acquisition of {[Y]}    & 1.00  & {[X]} 's acquisition of {[Y]}    & 0.95  &   acquisition of {[X]}  would reduce {[Y]} 's share  $^{\dagger}$   & 0.90 \\ 
{[X]} 's merger with {[Y]}    & 0.35  &    {[X]}  's merger with{[Y]}    & 1.00  & {[X]} 's merger with {[Y]}    & 0.94  &     {[X]} -  {[Y]} merger between  $^{\dagger}$    & 0.84  \\
{[X]} 's bid for {[Y]}    & 0.35  &    {[X]}  's bid for {[Y]}    & 1.00  & {[X]} 's bid for {[Y]}    & 0.97  &    part of {[X]} which  {[Y]} acquired $^{\dagger}$   & 0.83 \\ 
\hline
  \multicolumn{8}{c}{\bf founder-of}\\
\hline
{[X]} founder {[Y]}    & 0.68    &  {[X]} founder {[Y]}     & 1.00   & {[X]} founder {[Y]}      & 0.99   & founder of {[X]}  , {[Y]} $^{\dagger}$         & 0.97  \\
 {[X]}  CEO and founder {[Y]}   &    0.15       &   {[X]}  CEO and founder {[Y]}   &  1.00 &    {[X]}  CEO and founder {[Y]}   & 0.99  & co-founder of {[X]} 's millennial center , {[Y]}  $^{\dagger}$        & 0.94\\
{[X]}  's co-founder   {[Y]}   &   0.09 &  {[X]} owner {[Y]}     & 1.00  & {[X]} owner {[Y]}      & 1.00  & owned by {[X]} cofounder {[Y]}       & 0.95  \\
    &  &  {[X]} cofounder {[Y]}     & 1.00  & {[X]} cofounder {[Y]}      & 1.00  & Gates co-founded {[X]} with school friend {[Y]}    $^{\dagger}$      & 0.99  \\
    &    &  {[X]}  started by  {[Y]}     & 1.00  & {[X]} started by {[Y]}      & 1.00  &  who co-founded {[X]} with {[Y]}     $^{\dagger}$    & 0.95   \\
   &   &  {[X]} was founded by {[Y]}     & 1.00  & {[X]} was founded by {[Y]}      & 0.99   & to co-found {[X]} with partner {[Y]}     $^{\dagger}$     & 0.68  \\  
   &    &  {[X]} begun by {[Y]}     & 1.00  & {[X]} begun by {[Y]}      & 1.00  & {[X]} was started by {[Y]} , cofounder          & 0.98  \\
  &   &  {[X]} has established {[Y]}     & 1.00   & {[X]} has established  {[Y]}      & 0.99   &   set up   {[X]} with childhood friend  {[Y]}  $^{\dagger}$      & 0.96 \\
 &     &  {[X]}  chief executive and founder {[Y]}  & 1.00 &  {[X]}  co-founder and billionaire {[Y]}   $^{\ast}$        & 0.99  &    {[X]}  co-founder and billionaire {[Y]}  & 0.97  \\
\hline
  \multicolumn{8}{c}{\bf headquartered}\\
\hline
{[X]} headquarters in {[Y]}    & 0.95  &  {[X]}  headquarters in {[Y]}     & 1.00  & {[X]} headquarters in {[Y]}      & 0.98  & {[X]} headquarters in {[Y]}       & 0.98  \\
{[X]}  relocated its headquarters from {[Y]}    & 0.94   &  {[X]}  relocated its headquarters from {[Y]}        & 1.00   & {[X]}   relocated its headquarters from {[Y]}         & 0.98   & based at {[X]}  's suburban {[Y]} headquarters      $^{\dagger}$   & 0.98  \\
{[X]} head office in {[Y]}    & 0.84  &  {[X]} head office in {[Y]}     & 1.00  & {[X]} head office in {[Y]}      & 0.87  & head of {[X]} 's operations in {[Y]}  $^{\dagger}$      & 0.65   \\
{[X]} based  in  {[Y]}    & 0.75  &  {[X]} based in {[Y]}     & 1.00  & {[X]} based in {[Y]}      & 0.98  & branch of {[X]} company based in {[Y]}       & 0.98  \\
{[X]}  headquarters building in {[Y]}    & 0.67   &  {[X]}  headquarters building in {[Y]}        & 1.00   & {[X]}  headquarters building in {[Y]}         & 0.94   & {[X]} main campus in {[Y]}       & 0.99  \\
{[X]}  headquarters in downtown {[Y]}    & 0.64   &  {[X]}   headquarters in downtown {[Y]}        & 1.00   & {[X]}    headquarters in downtown {[Y]}         & 0.94   & {[X]}  headquarters in downtown {[Y]}       & 0.96  \\
{[X]} branch offices in {[Y]}    & 0.54  &  {[X]}  branch offices in  {[Y]}     & 1.00  & {[X]} branch offices in {[Y]}      & 0.98  & {[X]} 's {[Y]}  headquarters represented    $^{\dagger}$   & 0.98  \\
{[X]}  's corporate campus in {[Y]}    & 0.51   &  {[X]} 's corporate campus in {[Y]}        & 1.00   & {[X]} 's  corporate campus in {[Y]}         & 0.99   & {[X]}  main campus in {[Y]}       & 0.99  \\
{[X]}  's corporate office in {[Y]}    & 0.51   &  {[X]} 's corporate office in {[Y]}        & 1.00   & {[X]} 's  corporate office in {[Y]}         & 0.89   & {[X]} , {[Y]} 's corporate   $^{\dagger}$     & 0.94  \\
\hline
  \multicolumn{8}{c}{\bf affiliation}\\
\hline
{[X]} chief executive {[Y]}       & 0.92   &  {[X]}  chief executive  {[Y]}        & 1.00   & {[X]}  chief executive  {[Y]}         & 0.97   & {[X]}  chief executive {[Y]} resigned monday       & 0.94  \\
{[X]} secretary {[Y]}       & 0.88   &  {[X]}  secretary  {[Y]}        & 1.00   & {[X]}  secretary  {[Y]}         & 0.94   & worked with {[X]} manager {[Y]}       & 0.85  \\
{[X]} president {[Y]}       & 0.87   & {[X]} president {[Y]}       & 1.00   & {[X]} president {[Y]}       & 0.96  & {[X]} voted to retain {[Y]} as CEO   $^{\dagger}$     & 0.98   \\
{[X]} leader {[Y]}       & 0.72   &  {[X]}  leader  {[Y]}        & 1.00   & {[X]}  leader  {[Y]}         & 0.85   & head of {[X]} , {[Y]}     $^{\dagger}$    & 0.99   \\
{[X]} party leader {[Y]}       & 0.67   &  {[X]} party leader {[Y]}        & 1.00   & {[X]} party leader {[Y]}         & 0.87   &working with {[X]} , {[Y]} suggested    $^{\dagger}$     & 1.00  \\
{[X]} has appointed {[Y]}       & 0.63   &  {[X]} executive editor {[Y]}        & 1.00   & {[X]} has appointed {[Y]}         & 0.81   &{[X]}  president {[Y]}  was fired   & 0.90    \\
{[X]} player {[Y]}       & 0.38   &  {[X]}  player  {[Y]}        & 1.00   & {[X]}  player  {[Y]}         & 0.89   & {[X]} 's {[Y]} was fired    $^{\dagger}$     & 0.43   \\  
{[X]} 's secretary-general {[Y]}       & 0.36   & {[X]} 's secretary-general {[Y]}       & 1.00   & {[X]} 's secretary-general {[Y]}       & 0.93   &  Chairman of {[X]} , {[Y]}      $^{\dagger}$    & 0.88  \\
{[X]} hired {[Y]}       & 0.21   &  {[X]}  director  {[Y]}        & 1.00   & {[X]}  hired  {[Y]}         & 0.56   &  {[X]} hired {[Y]} as manager      $^{\dagger}$    & 0.85 
\end{tabular}
}
\caption{Subset of the non-noisy extractors (simplified) with their confidence scores $\mbox{cnf}(\lambda, {\cal G})$ learned in different configurations for each relation. ${\ast}$ denotes that the extractor was never learned in config\textsubscript{1} and config\textsubscript{5}. ${\dagger}$  indicates that the extractor was never learned in config\textsubscript{1},  config\textsubscript{5} and  config\textsubscript{9}. 
[X] and [Y]  indicate placeholders for entities.}
\label{top-patterns}
\end{table*}

\subsection{Deep Dive into Attributes of Extractors}
We analyze the extractors $\Lambda$ generated in BREE, BRET and BREJ for the 4 relations to demonstrate the impact of joint bootstrapping. 
Table~\ref{patternscores} shows the attributes of $\Lambda$. We manually annotate the extractors as {\it noisy} and {\it non-noisy}. 
We compute $ANNLC$ and the lower values in BREJ compared to BREE suggest fewer non-noisy extractors with lower confidence in BREJ due to the scaled confidence scores. 
{\it ANNE} (higher), {\it ANNLC} (lower), {\it AP} (higher) and {\it AN} (lower) collectively indicate that BRET mostly generates {\it NNHC} extractors. 
{\it AP} and {\it AN} indicate an average of $N_{+}(\lambda, {\cal G}_{l})$ (line `` (i)" \figref{brexconfig}) 
for positive and negative seeds, respectively for $\lambda \in \Lambda$ in the three systems. 
Observe the impact of scaling positive extractions ({\it AP}) in BREJ that shrink $\frac{N_+(\lambda,{\cal G}_n)}{N_+(\lambda,{\cal G}_p)}$ i.e., {\it ANP}.  
It facilitates $\lambda_{NNLC}$ to boost its confidence, i.e., $\lambda_{NNHC}$ in BREJ 
suggested by {\it AES} that results in higher $\#out$ and recall (Table~\ref{BRESH-results}, BREJ).

\subsection{Weighting Negatives Vs Scaling Positives}\label{negVsscaling}
As discussed, Table~\ref{BRESH-results} shows the performance of BREE, BRET and BREJ with the parameter $w_n=0.5$  in 
computing extractors' confidence $\mbox{cnf}(\lambda, {\cal G})$(\eqref{clusterconf}). 
In other words, config\textsubscript{9} (Table~\ref{BRESH-results}) is combination of both weighted negative and scaled positive extractions. 
However, we also investigate ignoring $w_n(=1.0)$ in order to demonstrate the capability of BREJ with only scaling positives and without weighting negatives. 
In Table~\ref{wnegtignored}, observe that BREJ outperformed both BREE and BRET for all the relationships due to higher $\#out$ and recall.  
In addition, BREJ scores are comparable to config\textsubscript{9} (Table~\ref{BRESH-results}) suggesting that the scaling in BREJ is 
capable enough to remove the parameter $w_n$. 
However, the combination of both weighting negatives and scaling positives results in the state-of-the-art performance.  


\subsection{Qualitative Inspection of Extractors}
Table~\ref{top-patterns} lists some of the non-noisy extractors (simplified) learned 
in different configurations to illustrate boosting extractor confidence $\mbox{cnf}(\lambda, {\cal G})$. 
Since, an extractor $\lambda$ is a cluster of instances, therefore to simplify, we show one instance (mostly populated) from every $\lambda$. 
Each cell in Table~\ref{top-patterns} represents either a simplified representation of $\lambda$ or its confidence.  
We demonstrate how the confidence score of a non-noisy extractor in BREE (config\textsubscript{1}) is increased in BREJ (config\textsubscript{9} and config\textsubscript{10}).  
For instance, for the relation {\it acquired}, an extractor \{{\it [X]  acquiring [Y]}\} is generated by BREE, BRET and BREJ; however, its confidence is 
boosted from $0.75$ in BREE (config\textsubscript{1}) to $0.95$ in BREJ (config\textsubscript{9}). Observe that BRET generates high confidence extractors. 
We also show extractors (marked by ${\dagger}$) learned by BREJ with $\mbox{sim}_{cc}$ (config\textsubscript{10}) but 
not by config\textsubscript{1}, config\textsubscript{5} and config\textsubscript{9}. 

\subsection{Entity Pairs: Ordered Vs Bi-Set}
In Table~\ref{BRESH-results}, we use ordered pairs of typed entities. Additionally, we also investigate using entity sets   
and observe improved recall due to higher $\#out$ in both BREE and BREJ, comparing correspondingly Table~\ref{entitybiset} 
and \ref{BRESH-results} ({\it baseline} and config\textsubscript{9}). 

\section {Conclusion}
We have proposed a Joint Bootstrapping Machine for relation extraction (BREJ) 
that takes advantage of both entity-pair-centered and template-centered approaches. 
We have demonstrated that the joint approach scales up positive instances that boosts the confidence of NNLC extractors and improves recall.  
The experiments  showed that the cross-context similarity measures improved recall and 
suggest removing in total four parameters.

\begin{table}[t]
\centering
\def\arraystretch{1.17}
\resizebox{0.46\textwidth}{!}{
\begin{tabular}{r|cccc|cccc}
\multirow{2}{*}{\it Relationships} & \multicolumn{4}{c|}{{\bf BREE} + $\mbox{sim}\dnrm{match}$}  & \multicolumn{4}{c}{{\bf BREJ} + $\mbox{sim}\dnrm{match}$} \\ \cline{2-9}
 &$\#out$ & $P$    & $R$           & $F1$   &$\#out$ & $P$    & $R$           & $F1$   \\
\hline
acquired          & 2786     & .90  &  \underline{.50}       &   \underline{.64}     & 21733     & .80  &  \underline{.87}       &   .83       \\
founder-of        & 543     &  1.0  &  .67       &   .80     & 31890     & .80  &  .99       &   .89      \\
headquartered       & 16832     & .62  &  \underline{.81}       &   \underline{.70}   & 52286     & .64  &  \underline{.94}       &   .76      \\  
affiliation   & 21812     & .99  &  \underline{.74}       &   \underline{.85}     & 42601     & .96  &  \underline{.85}       &   \underline{.90}\\ \hline
{\it avg}    & 10493     & .88  &  .68       &   \underline{.75}     & 37127     & .80   &  .91       &   .85 
\end{tabular}}
\caption{BREX+sim\dnrm{match}:Scores with entity {\it bisets}}
\label{entitybiset}
\end{table}



\section*{Acknowledgments}
We thank our colleagues  Bernt Andrassy,  Mark Buckley,  Stefan Langer, Ulli Waltinger and Usama Yaseen, and 
anonymous reviewers for their review comments. 
This research was supported by Bundeswirtschaftsministerium ({\tt bmwi.de}), grant 01MD15010A (Smart Data Web) 
at Siemens AG- CT Machine Intelligence, Munich Germany.


\bibliography{acl2017}
\bibliographystyle{acl_natbib}

\end{document}